\documentclass[12pt]{article}
\usepackage{newtxtext,newtxmath} 
\usepackage{PRIMEarxiv}
\usepackage{graphicx} 
\usepackage{subcaption} 
\usepackage{enumitem}
\usepackage{comment}
\usepackage{amsmath, amsfonts}
\usepackage{hyperref}
\usepackage{url}
\usepackage{booktabs}
\usepackage{nicefrac}
\usepackage{microtype}
\usepackage{lipsum}
\usepackage{svg}
\usepackage{parskip}
\usepackage{multirow}
\definecolor{lightgreen}{rgb}{0.56, 0.93, 0.56}
\definecolor{lightred}{rgb}{0.98, 0.81, 0.81}
\definecolor{mygreen}{rgb}{0.0, 0.5, 0.0}
\definecolor{myred}{rgb}{0.9, 0.0, 0.0}
\definecolor{myblue}{rgb}{0.8, 0.8, 1.0}
\usepackage{colortbl}
\usepackage{caption} 
\usepackage{float} 

\usepackage{dirtytalk}

\usepackage{geometry}
\graphicspath{{media/}} 

\title{Assessment of GPT-4's performance in a \\ USMLE-based case study}

\author{
     \textbf{Uttam Dhakal}$^{1,\dag}$,
    \textbf{Aniket Kumar Singh}$^{2,\dag}$,
    \textbf{Suman Devkota}$^{1,\dag}$, \\
    \textbf{Yogesh Sapkota}$^{1,\dag}$,
    \textbf{Bishal Lamichhane}$^{3, \dag}$,
    \textbf{Suprinsa Paudyal}$^{4}$,
    \textbf{Chandra Dhakal}$^{5,\*}$
    \\
    $^{1}$Department of Electrical and Computer Engineering, Youngstown State University, OH, 44555, U.S.A\\
    $^{2}$Department of Computing and Information Systems, Youngstown State University, OH, 44555, U.S.A\\
    $^{3}$Department of Mathematics and Statistics, University of Nevada, Reno, Nevada, 89557, USA\\
    $^{4}$Department of Chemical and Biological Sciences, Youngstown State University, OH, 44555, U.S.A\\
    $^{5}$Formerly with the Department of Agricultural and Applied Economics, University of Georgia, Athens, GA 30602, U.S.A\\
    $^\dag$These authors contributed equally to this work.\\
$^*$Correspondence: \texttt{chandra.dhakal25@uga.edu} \\
}

\begin{document}
\maketitle

\begin{abstract}

This study assess the performance of GPT-4 in healthcare applications. A simple prompting technique was used to prompt the Large Language Model (LLM) with questions taken from the United States Medical Licensing Examination (USMLE) questionnaire and it was tasked to evaluate its confidence score before posing the question and after asking the question. The questionnaire was categorized into two groups-questions with feedback (WF) and questions with no feedback(NF) post-question. The model was asked to provide absolute and relative confidence scores before and after each question. The experimental findings were analyzed using statistical tools to study the variability of confidence in WF and NF groups. Additionally, a sequential analysis was conducted to observe the performance variation for the WF and NF groups. Results indicate that feedback influences relative confidence but doesn't consistently increase or decrease it. Understanding the performance of LLM is paramount in exploring its utility in sensitive areas like healthcare. This study contributes to the ongoing discourse on the reliability of AI, particularly of LLMs like GPT-4, within healthcare, offering insights into how feedback mechanisms might be optimized to enhance AI-assisted medical education and decision support.

\end{abstract}


\section{Introduction}

Large Language Models (LLMs) have demonstrated impressive performance in the various sectors\cite{min2023recent}. With advanced tuning capabilities incorporated into models such as GPT-4 and GPT-3.5, they surpass human capabilities in areas like programming, policy making, and solving complex problems\cite{gokul2023llms}. LLMs contribute significantly to Artificial Intelligence (AI) by understanding natural language and producing extensively accurate natural language with minimal human intervention\cite{zhuang2023efficiently}. They find applications in various AI contexts, including document management\cite{kasneci2023chatgpt}, chatbots\cite{flanagin2023guidance}, text-based games\cite{yao2023tree}, law\cite{biswas2023role}, human resources\cite{budhwar2023human}, knowledge extraction\cite{hong2023knowledge}, market research\cite{saputra2023impact}, and healthcare\cite{johnson2023assessing}.
LLMs exhibit considerable strength and versatility for various applications, so a comprehensive understanding of their operation becomes important. It is paramount to understand the operation of these models. In critical domains such as healthcare, policy-making, and law enforcement, where minimizing risk is imperative due to the potential for significant losses from erroneous decisions, it is crucial to gain insight into the functionality of these models. 

 Our previous efforts explore the potential of conducting such studies and underscore their importance \cite{singh2023confidencecompetence}.
Especially in the health domain, understanding how these models operate and their cognitive abilities is crucial. With the current state-of-the-art LLMs, they can diagnose a disease based on symptoms and patient information  \cite{Holohan_2023}. Our study aims to explore and analyze the intricate interactions within the current capabilities of LLM, specifically focusing on GPT-4 when it comes to dealing with health-related data.  To evaluate the performance of GPT-4, we evaluated using one hundred sets of questions from the United States Medical Licensing Examination (USMLE). The USMLE is a three-step examination required to obtain a medical license in the United States. It comprises questions that assess basic science knowledge, clinical knowledge, and clinical skills. The exam evaluates an individual's understanding and ability to apply medical knowledge.

Based on our findings, it is evident that these advanced language model display a remarkable level of confidence and accuracy in addressing medical inquiries. Moreover, their confidence levels significantly rose when providing accurate responses. Notably, in certain cases, a similar boost in confidence was observed even when delivering incorrect answers. The model exhibited reduced confidence, even when delivering correct answers, especially in scenarios where feedback was incorporated. When incorporating feedback (WF), the model's confidence level was around 0.9, and it recorded an 88\% accuracy.
In contrast, without feedback (NF), the model's confidence increased to around 0.95, achieving a higher accuracy of 92\%. The behaviors of LLMs should be carefully considered before employing them in application, especially in critical fields like healthcare. This study sheds light on the potential of LLMs, particularly GPT-4, to effectively handle health-related data, thereby paving the way for further advancements in the field.

\section{Related works}

The growing use of LLMs in the healthcare domain has piqued the interest of researchers worldwide, leading to numerous published studies. A plethora of noteworthy research studies, spanning from AI-generated medical advice \cite{haupt2023ai}, biomedical text generation \cite{luo2022biogpt}, language models for radiology \cite{liu2023radiology}, and models for health care services \cite{javaid2023chatgpt}. 

Numerous studies investigating the integration of ChatGPT and the USMLE have been conducted. The study by Brin et al. \cite{brin2023comparing} compared the performance of ChatGPT and GPT-4 in the context of USMLE soft skill assessments. The results revealed that GPT-4 exhibited superior performance by correctly answering 90\% of USMLE soft skills questions compared to ChatGPT's 62.5\%. A similar study by Nori et al. \cite{nori2023capabilities} not only compared ChatGPT and GPT-4 but also compared another model called Flan-PaLM and concluded that GPT-4 significantly outperformed other models. The study's findings conclusively demonstrated that GPT-4 exhibited significant superiority over the other models under consideration. Patel et al. \cite{patel2023limits} evaluated the impact of different prompt engineering strategies on ChatGPT in addressing medical problems, specifically focusing on medical calculations and clinical scenarios. 

\subsection{Cognitive Abilities of LLMs}

Studies have shown that LLMs possess capabilities that align with certain cognitive abilities traditionally associated with humans. These include chain-of-thought reasoning and aspects of the theory of mind, allowing them to generate coherent and contextually appropriate text outputs. Their proficiency in tasks requiring abstract knowledge and reasoning has characterized them as \say{thinking machines} and raised questions about their potential cognitive capacities, which may extend to goal direction, agency, and executive  \cite{zhuang2023efficiently}\cite{Lamichhane2023Nuances}.

The GPT series, as an example of LLMs, are trained on vast text corpora with the primary objective of predicting the next or missing words in sentences. This training has evolved to include additional objectives to enhance their language processing capabilities \cite{pellert2022ai}. Emerging studies suggest that LLMs may exhibit cognitive abilities, but systematic evaluation is necessary to substantiate these claims. This involves multiple tasks, control conditions, iterations, and statistical robustness tests. Moreover, when fine-tuned with data from psychological experiments, LLMs have mirrored human behavior and, in some cases, outperformed traditional cognitive models, particularly in decision-making domains \cite{bill2023fine}.

\subsection{AI Ethics in Healthcare}
The integration of AI into healthcare brings forth significant ethical challenges. Informed consent, safety and transparency, algorithmic fairness and biases, and data privacy are among the primary concerns that require thorough examination. As AI reshapes clinicians' roles and practices, ethical frameworks are crucial to guide the integration of AI in clinical decision-making. Best practices developed by leading institutions aim to address transparency, fairness, and privacy concerns, promoting ethical AI implementation in healthcare \cite{jobin2019global}. A systematic approach is necessary to identify and address gaps in the ethical application of AI, paving the way for evidence-informed practices.
The applications of AI in healthcare underscore the necessity of accurate and ethically sound decision-making \cite{lysaght2019ai}. As LLMs are increasingly adopted for tasks such as medical diagnosis, the confidence-competence gap, and its implications have become a critical focus of AI safety and reliability. Current AI in healthcare literature suggests a need for rigorous standards and ethical oversight to ensure that the deployment of AI does not compromise patient safety or care quality.

 Simple prompting techniques are exclusively employed in this study to elicit responses from GPT-4 on clinical knowledge sets derived from USMLE questions. The accuracy of responses provided by GPT-4 is assessed, with the dataset comprising instances of both correct and incorrect responses generated by ChatGPT, along with the associated confidence levels. The study is conducted under two distinct settings: one where no feedback (NF) is provided to GPT-4 and another where feedback (WF) is incorporated, explicitly informing the model about the correctness of its responses.

 \section{Methodology}
 
In this section, the approach of data organization and analyses are detailed. The primary tool for assessment was the GPT-4 model, and the questionnaire comprised questions predominantly from the USMLE question bank. The LLM model's proficiency in healthcare knowledge and ability to assess confidence has been evaluated based on its response to the questionnaire. In addition, 16 questions for the questionnaire were taken from high school biology textbooks. While these questions may not have the rigor of the USMLE test, these questions were interspersed randomly in the questionnaire. One of the areas of interest of this study was to observe how the models assessed and adjusted their confidence level between random questions of varying difficulties when transitioning between questions of varying difficulties.  Assessing top-tier models like GPT-4 is critical, especially when considering their implication in sensitive and vital areas such as healthcare. Understanding their capabilities and limitations is crucial before considering their deployment in such critical sectors.

\subsection{Data Collection} 
GPT-4 was asked multiple-choice questions from the USMLE question pool to gather data for this study. These questions covered a wide range of topics with varying levels of difficulty. The difficulty levels of the questions were not disclosed to the model. Some questions were challenging, based on the USMLE standards, while others were easier, focusing on general health topics. A simple prompting strategy was employed to ensure that the model accurately comprehended the questions.

\begin{figure}[ht!]
    \centering
    \begin{subfigure}[b]{0.342\textwidth}
        \includegraphics[width=\textwidth]{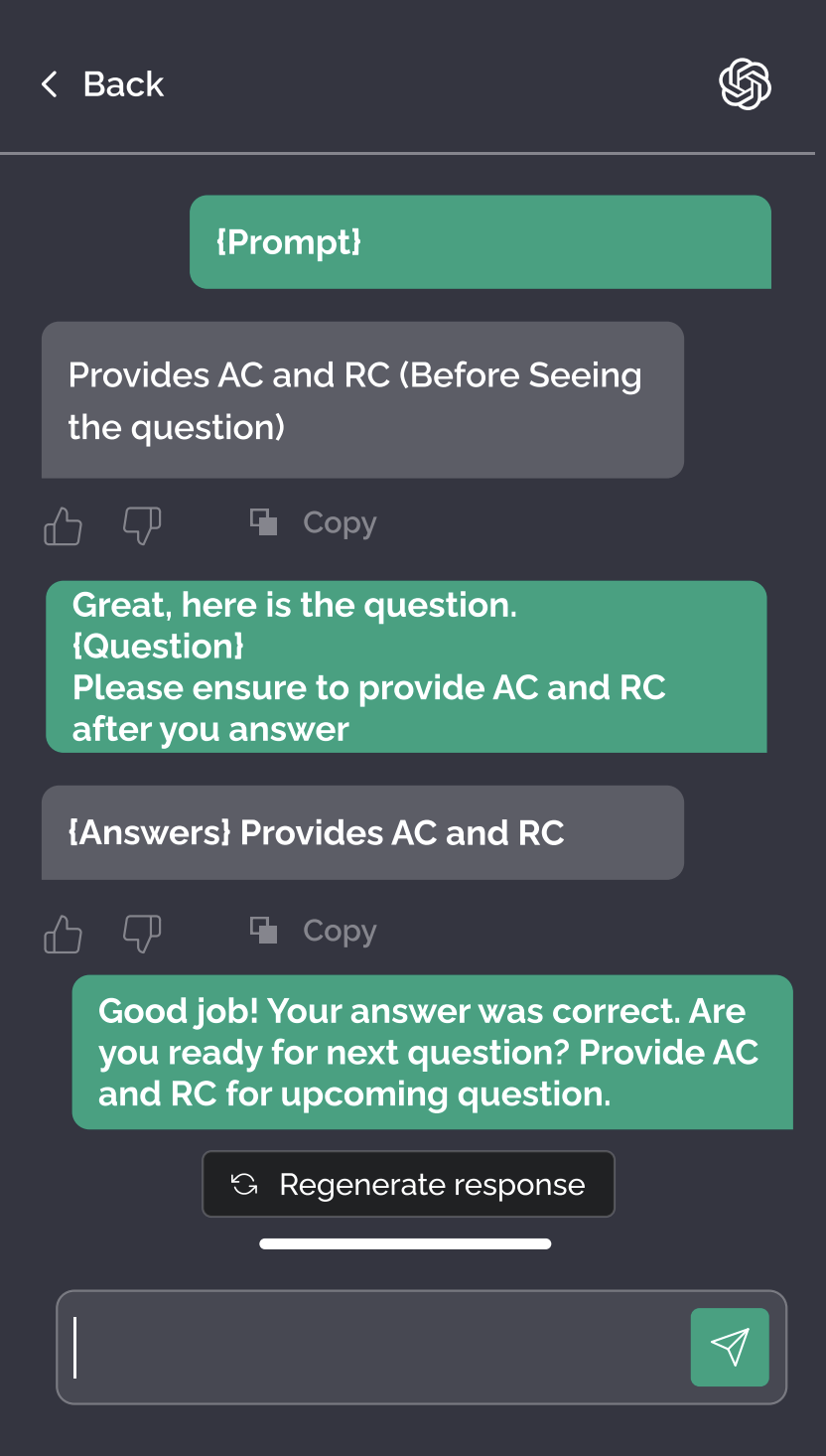}
        \label{fig:chat_feedback}
    \end{subfigure}
    \hspace{1cm}
    \begin{subfigure}[b]{0.35\textwidth}
        \includegraphics[width=\textwidth]{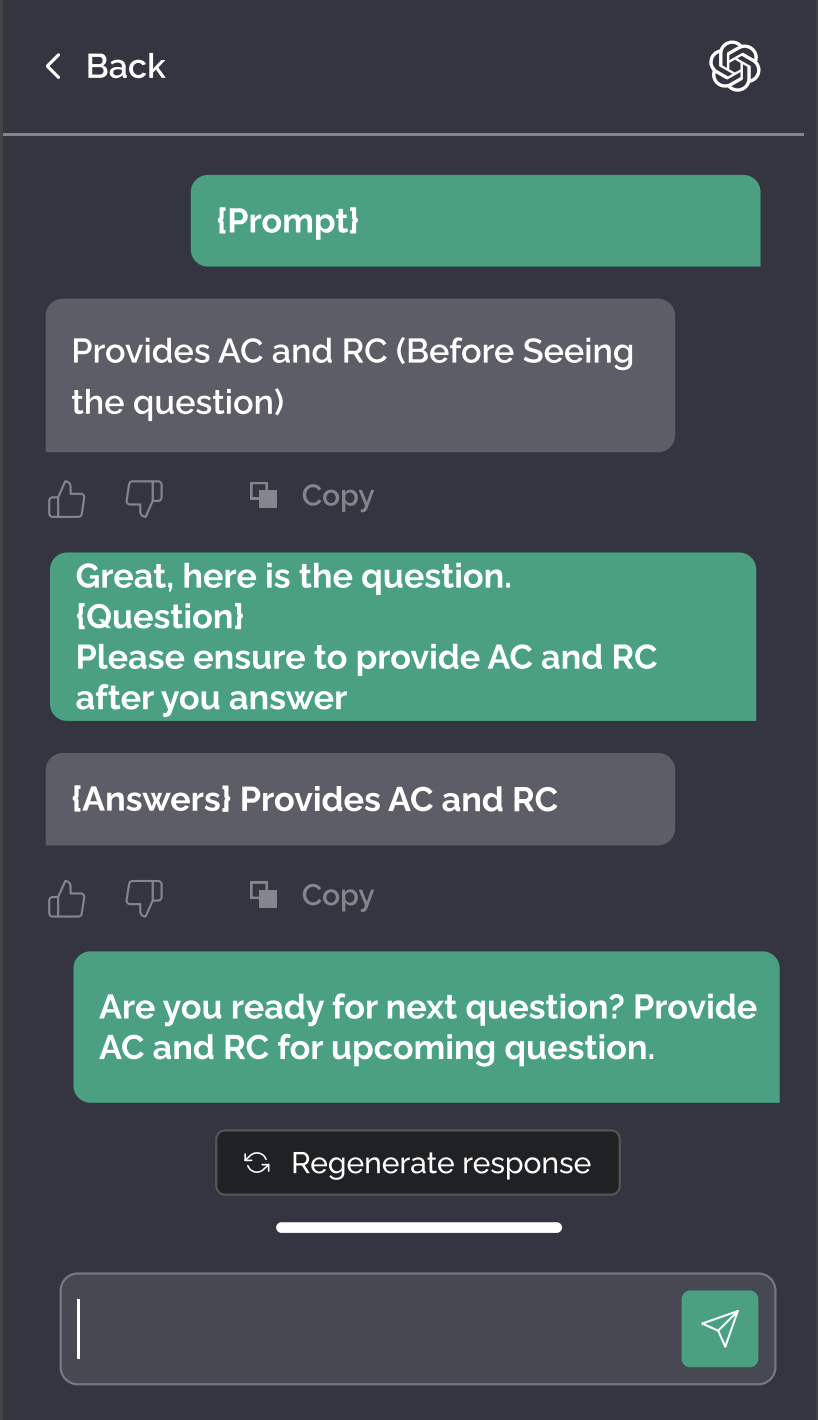}
        \label{chat_nofeedback}
    \end{subfigure}
    \caption{Interaction with ChatGPT.  On the left with Feedback, and the right without feedback. Actual conversation is shown in Figure \ref{chat_interface} in Appendix \ref{appendix_data}.}
    \label{chatinterface}
\end{figure}

Figure \ref{chatinterface} shows the interaction process with ChatGPT.  Before asking the questions, the model was instructed to evaluate and quantify its confidence levels, employing a probabilistic scale for assessment. It involved LLM providing Absolute Confidence (AC) and Relative Confidence (RC) values, with the scale ranging from 0, indicating no confidence, to 1, signifying full confidence. This preemptive confidence assessment was documented before the question was posed. Subsequently, the language model was tasked with reevaluating its confidence levels after responding to the presented question, and the process was repeated for each question. The inquiry involved soliciting the models' self-assessment of their proficiency in addressing the posed question. The analysis of these self-ratings aimed to clarify the correlation between the models' perceived confidence and their actual performance. While asking questions to the model, we focused on ensuring the prompts were the same and the questions were clear. A \textit{Simple Prompting Technique} is used in our study. The GPT-4 model is far more advanced than most currently available LLMs and can comprehend the test using just simple prompting\cite{singh2023confidencecompetence} \cite{info15020092}.

\begin{quote}
    \textbf{Simple prompt with explanation} \\
    \textit{In this exercise, we're exploring the self-awareness of the Large Language Model (LLM). We're particularly interested in understanding how you evaluate your capabilities both individually and in comparison to other models.}\\
   \textit{ Please rate your confidence on a scale from 0 (not confident) to 1 (confident):}
    \begin{enumerate}[label=\alph*)]
        \item \textit{How confident are you in answering the upcoming questions? (Absolute Confidence)}
        \item \textit{Compared to other LLMs, how confident are you in your ability to answer the questions? (Relative Confidence)}
    \end{enumerate}
  \textit{  After completing the questions, we'll measure your confidence levels again.}
\end{quote}

A 0-1 confidence score scale was implemented to gauge the model's confidence, where 0 represents the lowest confidence score and 1 being the highest confidence score. The model was asked to provide its answers and a confidence score before and after viewing each question as shown in Figure \ref{chatinterface}. This approach enabled us to record the model's initial confidence (before seeing the question) and their revised confidence (after seeing the question), with a specific focus on factors such as AC1 (Absolute Confidence pre-viewing), RC1 (Relative Confidence pre-viewing), AC2 (Absolute Confidence post-viewing), and RC2 (Relative Confidence post-viewing). This methodology helped us understand the accuracy of the model's answers and how their perceived confidence in their answers changed upon reviewing the questions. By analyzing these factors, insights into the model's self-awareness of their abilities and adaptability in adjusting confidence levels based on question difficulty were obtained.

\subsubsection{With Feedback vs. No Feedback} Two groups,  WF and NF, were employed to gather data for assessing the self-evaluation of a language model. The feedback mechanism aimed to streamline confidence and self-assessment to understand the effect of feedback in enhancing confidence levels. This process involved providing feedback after posing a question to the model. Specifically,  the model was asked to rate its absolute and relative confidence before and after answering the question. Additionally, before rating absolute and relative confidence, the model was informed about the correctness of its response. The objective was to evaluate the effects of the WF and NF group on the model's self-assessment. For the NF group, the model was not purposely informed about the accuracy of its response.  

\subsection{Question Design}
The USMLE exam is one of the toughest tests taken globally\cite{Erudera_2023} and is well--recognized in the medical field. Since USMLE exam questions are the standards in assessing the proficiency of medical students, they choose to assess knowledge of the LLM as well in addition to a few questions from the high school level. These questionnaires often present real-life scenarios, including details like a patient's age, personal traits, medical history, habits, and other factors affecting their health. The exam questions demonstrate thoughtful construction to thoroughly test the critical competencies expected of medical graduates who will enter clinical practice.
There is broad coverage of topics across all major organ systems, spanning various diseases, patient demographics, and clinical settings. This requires examines to have a strong generalist medical knowledge foundation. Questions range from foundational concepts like anatomy and physiology to complex clinical reasoning about diagnosis and management. The vignette-based format presents a short clinical scenario with details like patient age, symptoms, medical history, and initial workup findings. This emulates real-world clinical encounters and tests the application of knowledge, not just memorized facts. The vignettes provide sufficient patient details, so the information is neither vague nor an unrealistic laundry list. This makes the task clear to the examinee yet still challenging. The questions avoid ambiguous phrasing and indicate what is being asked. The answer options are thoughtful and well-differentiated. The correct answer is clear, concise, and well-supported by the vignette. Incorrect options include common mistakes and misconceptions a novice might make rather than completely irrelevant answers. This appropriately distinguishes between superficial knowledge and true mastery. The questions vary in difficulty level, including some testing core concepts and others requiring deep analysis of clinical findings or scientific principles. This discriminates the depth of knowledge across candidates. 


\subsection{Variable Definitions}
During the data collection process, we employed absolute and relative confidence. Absolute confidence refers to the level of certainty or confidence the model has in its answer to a question, irrespective of external factors. It represents the model's independent evaluation of how confident it feels about its response. Relative confidence, on the other hand, compares the model's confidence in its answer to the question with its confidence in answers across different questions compared to other language models. AC1 and RC1 represented the model's absolute and relative confidence levels before the model questions were posed to the model. Whereas, AC2 and RC2 denote the absolute and relative confidence levels after posing the question to the model for each instance. This approach was consistent across both the With feedback (WF) and No feedback (NF) groups. 

\section{Results}
 The result section encompasses descriptive statistics, correlation analysis, comparative examination of feedback groups, and visualization. Across all instances, the median confidence scores for AC1, AC2, RC1, and RC2 consistently reached 0.9.  Notably, the WF group achieved a response rate of 88 questions answered correctly out of 100 questions, whereas the NF group demonstrated a slightly higher accuracy, with 92 out of 100 questions answered correctly. The model showed varied confidence depending on whether feedback is provided or not. The mean confidence for AC1 before asking a question was 0.91, and for RC1, it was 0.9. After posing the question, the mean for AC increased to 0.94, and for RC, it rose to 0.93, indicating a higher confidence after asking the question. The standard deviation was higher for AC1 at 0.05 and lower for RC2 at 0.046, suggesting greater variability in AC1. AC1 and AC2 ranged from 0.7 to 1, whereas RC1 and RC2 ranged from 0.75 to 1, showing different confidence levels. The mean values for AC1 and AC2 closely matched the overall mean for the WF condition. After the feedback, the standard deviation for AC1 and AC2 decreased, whereas it increased for RC1 and RC2. The minimum values for AC1, AC2, RC1, and RC2 were 0.7, 0.78, 0.76, and 0.81 respectively with the maximum being 1. For the NF group, mean values for AC1 and AC2 were similar. The standard deviation for AC1 and AC2 was higher than the overall standard deviation, while for RC1 and RC2, it was lower. The minimum confidence values for AC1 and AC2 were 0.7, and for RC1 and RC2, they were 0.75. The maximum values for AC1, AC2, RC1, and RC2 were observed to be 0.98, 0.99, and 1 respectively.

\begin{figure}[ht]
    \centering
    \includegraphics[width=0.8\textwidth]{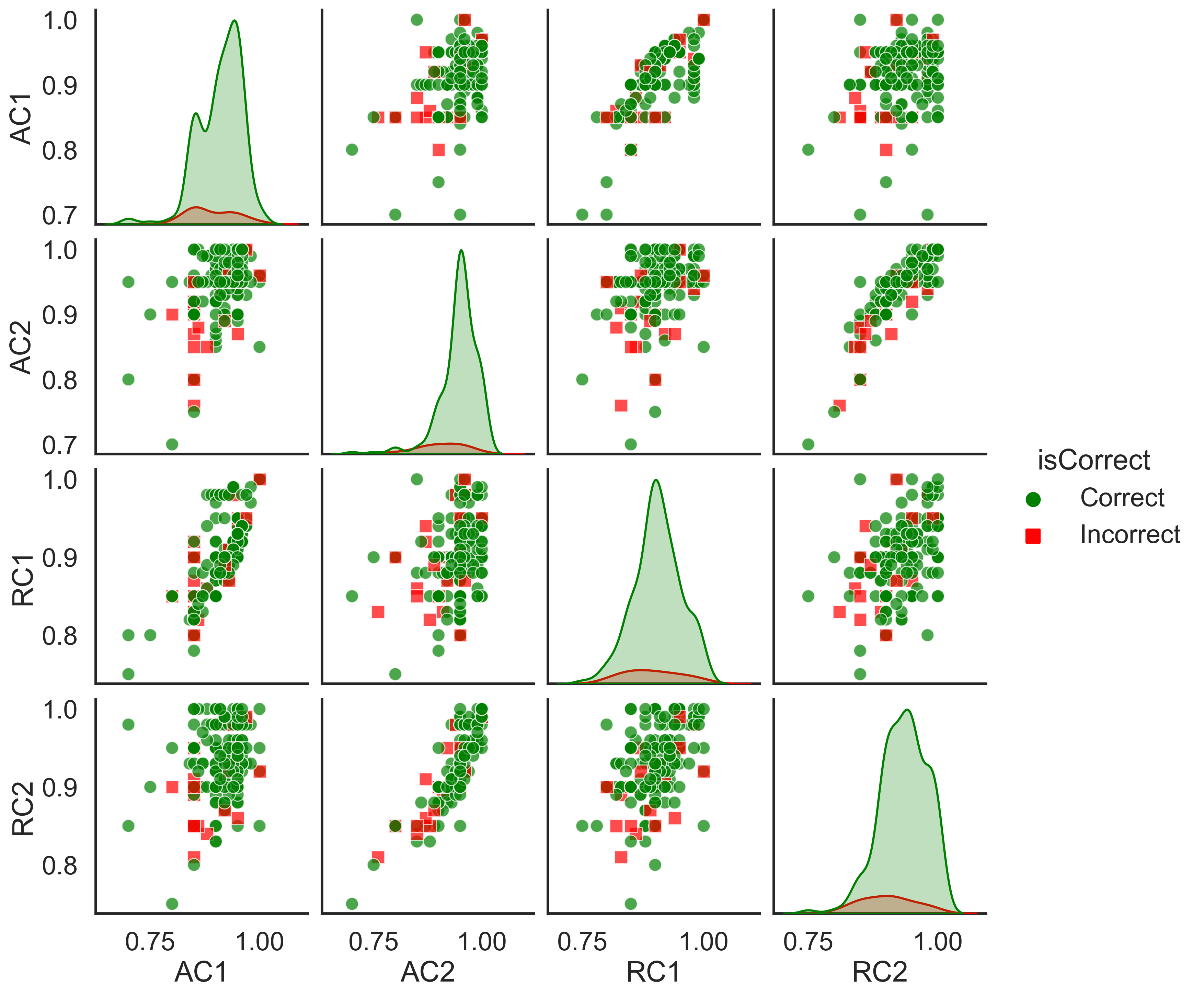}
    \caption{Pair Plot AC vs RC}
    \label{fig:pairPlot}
\end{figure}

\subsection{Visualizations}
\begin{figure}[ht]
    \centering
    \includegraphics[width=0.9\textwidth]{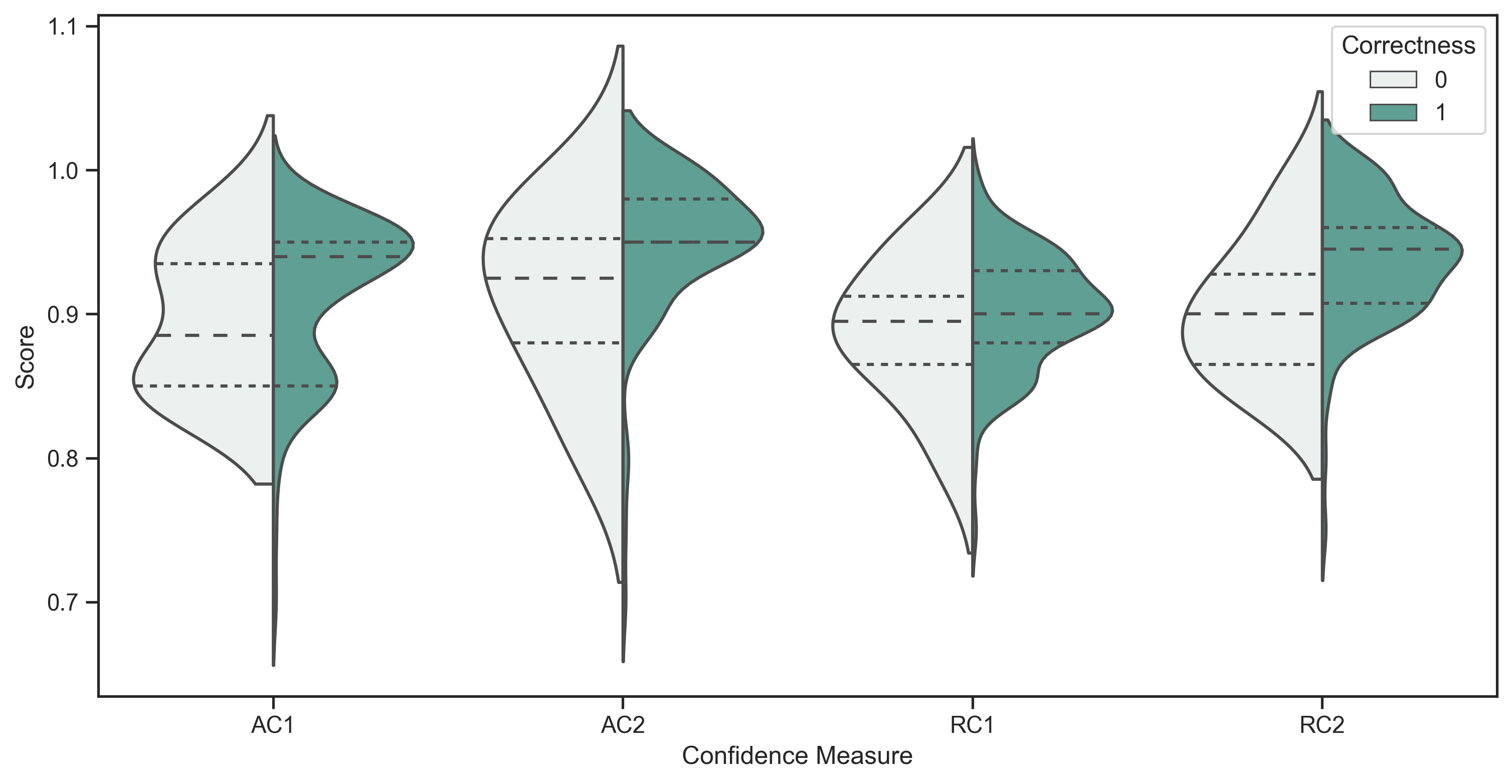}
    \caption{Violin Plot for No Feedback group}
    \label{fig:no_feedback_group}
\end{figure}

The pair plot in Figure \ref{fig:pairPlot} illustrates the correlation between absolute and relative confidence in the model. The model exhibited higher confidence levels of more than 0.8 in most cases when providing accurate responses. Moreover, the model showed higher confidence levels even when responding incorrectly. AC1 consistently showed higher values of around 0.85 before posing questions, whereas AC2 maintained or increased value after responding accurately. There were instances where the model showed lower confidence despite responding accurately. Additionally, the model also showed lower confidence for incorrect answers. \\
The model's performance is depicted in violin plot Figure \ref{fig:no_feedback_group} for the NF group. For AC1, the model's median confidence score is below 0.9 for wrong answers with confidence values spanning from 0.87 to 1. For correct answers, LLM median confidence score was above 0.9 with a confidence score between 0.78 and 1.

\begin{figure}[H]
    \centering
    \includegraphics[width=0.9\textwidth]{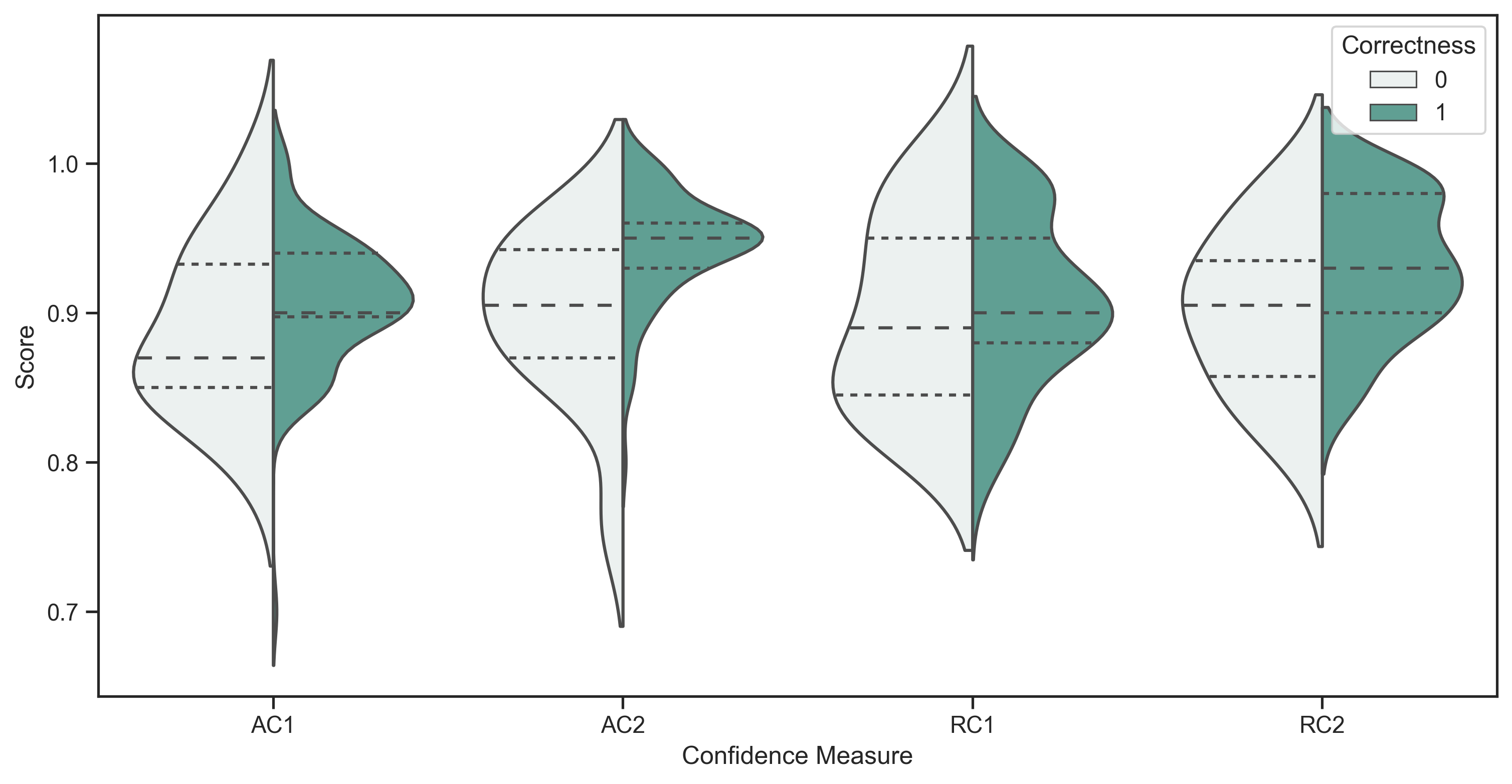}
    \caption{Violin Plot for Feedback group}
    \label{fig:feedback_group}
\end{figure}

The violin plot for the feedback group is depicted in \ref{fig:feedback_group}. For RC1, the confidence level showed no fluctuation for both the WF and NF groups. The model showed greater confidence for correct answers compared to incorrect ones for RC2. RC remains unchanged compared to AC. For the WF group, AC1 and AC2 as well as RC1 and RC2 reached the maximum.

\begin{figure}[ht]
    \centering
    \begin{subfigure}[b]{0.49\textwidth}
        \includegraphics[width=\textwidth]{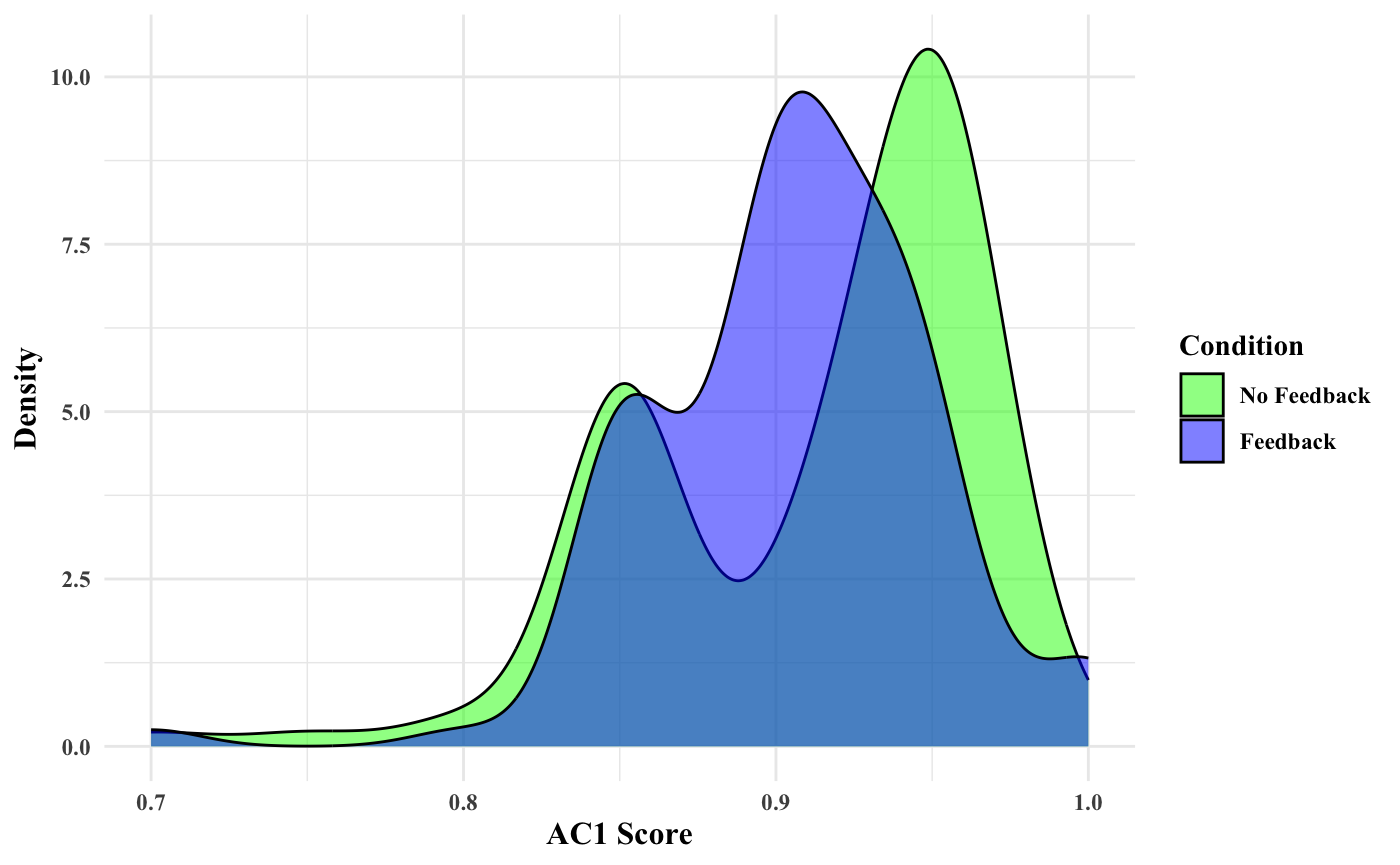}
        \label{fig:ac2_1}
    \end{subfigure}
    \hfill 
    \begin{subfigure}[b]{0.49\textwidth}
        \includegraphics[width=\textwidth]{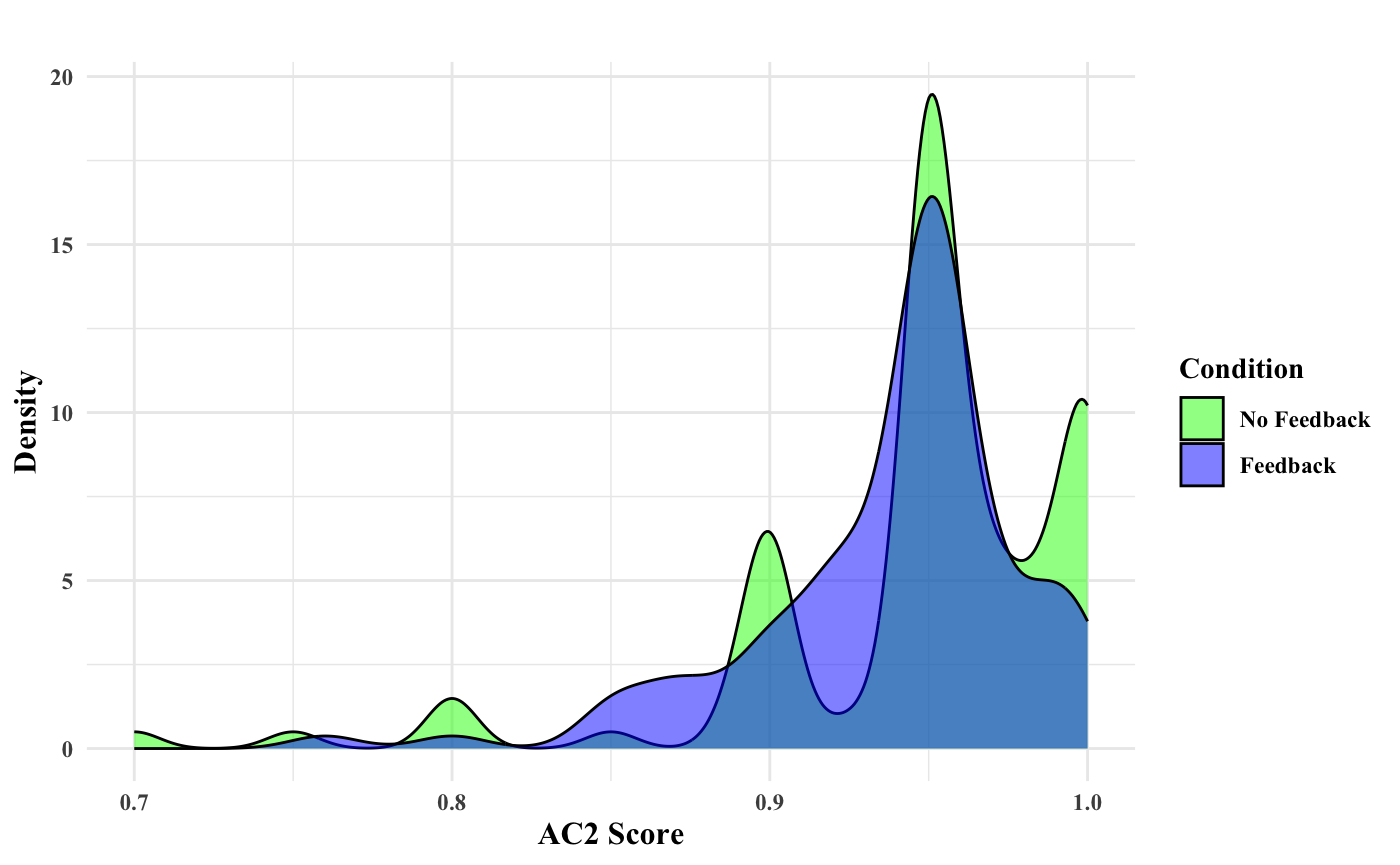}
        \label{fig:ac2_2}
    \end{subfigure}
    \caption{Distribution plot for AC1 and AC2}
    \label{fig:ac1_distributions}
\end{figure}

The distribution plot is shown in Figure. \ref{fig:ac1_distributions} presents a density plot comparing AC1 scores for both groups--WF(blue) and NF(green). The x-axis displays the AC1 score, ranging approximately from 0.7 to 1, indicating a measure of confidence or performance, while the y-axis shows the density of these scores. The distribution plot depicts the impact of feedback on the AC1 performance of the model. The AC1 scores for the WF and NF groups differ, with a higher AC1 score for prompts in the WF group. \newline
Although there is no notable difference between the WF and NF groups, it is observed that the absolute confidence before posing the question was higher at around 0.95. After receiving the feedback, the confidence score was around 0.9. \newline
It indicates that the model may either over-adjust in response to feedback, leading to errors, or re-calibrate its responses to maintain consistency. From this observation, it is not easy to determine the impact of feedback conclusively. However, feedback has a significant influence on the model's performance. 

It is important to understand the model's ability to assess its confidence for both WF and NF cases. Figure \ref{fig:ac1_distributions} shows the distribution of AC1 and AC2 scores for both WF and NF groups. AC1 score shows greater variability for the NF group compared to the WF group. There are instances where the language model's confidence scores are higher in the absence of feedback. However, statistical tests were employed to understand the significance of this observed confidence score difference. The normality assumption is not satisfied for the AC2 score. Thus, a non-parametric Wilcoxon Signed-Rank test was employed. AC2 scores for the WF and the NF groups were found to have no significant statistical difference. Null hypothesis failed for all four tests (AC1, AC2, RC1, RC2) which concluded that the scores difference is not significant for both WF and NF groups.

\begin{figure}[ht]
    \centering
    \begin{subfigure}[b]{0.49\textwidth}
        \includegraphics[width=\textwidth]{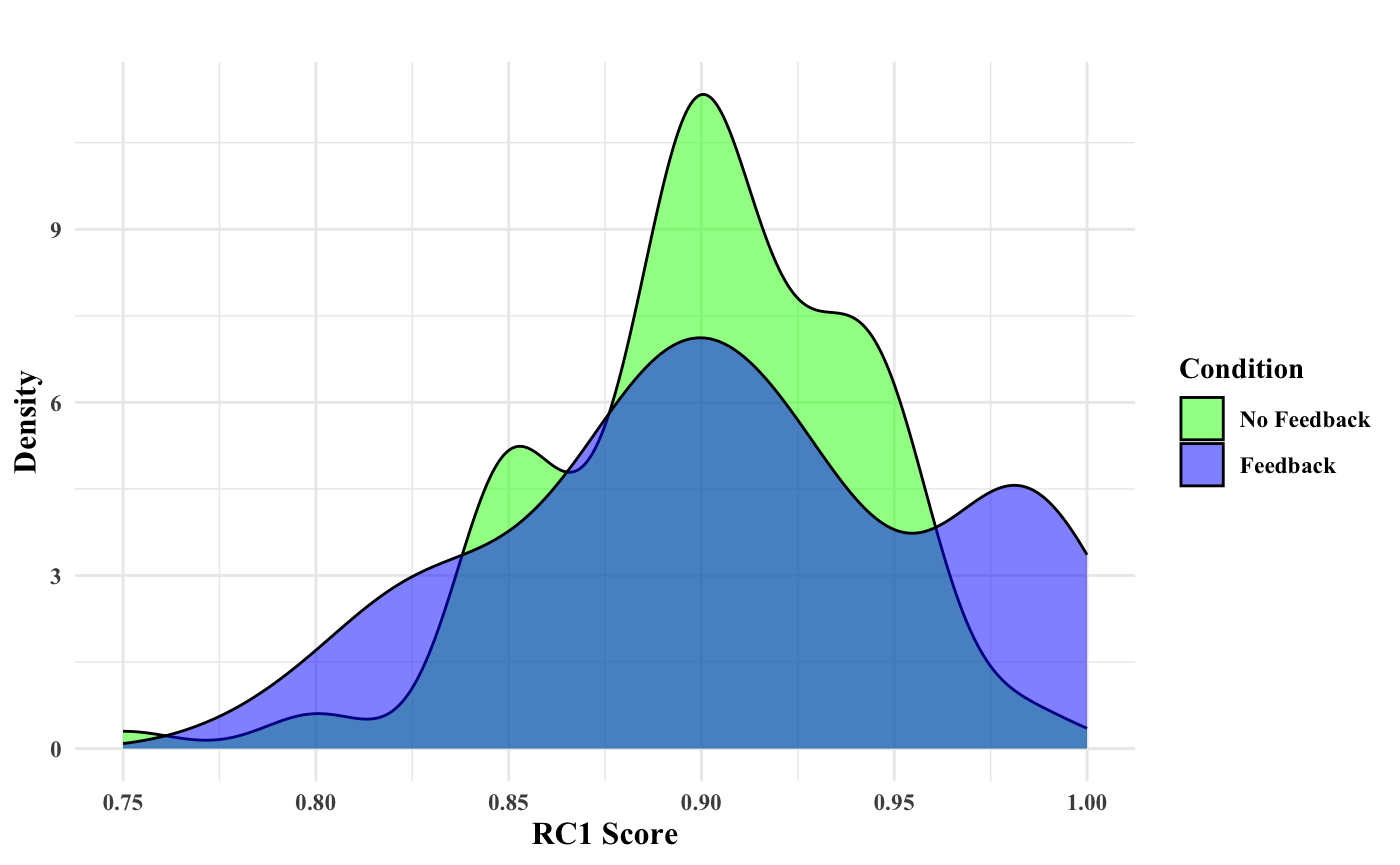}
        \label{fig:rc2_1}
    \end{subfigure}
    \hfill 
    \begin{subfigure}[b]{0.49\textwidth}
        \includegraphics[width=\textwidth]{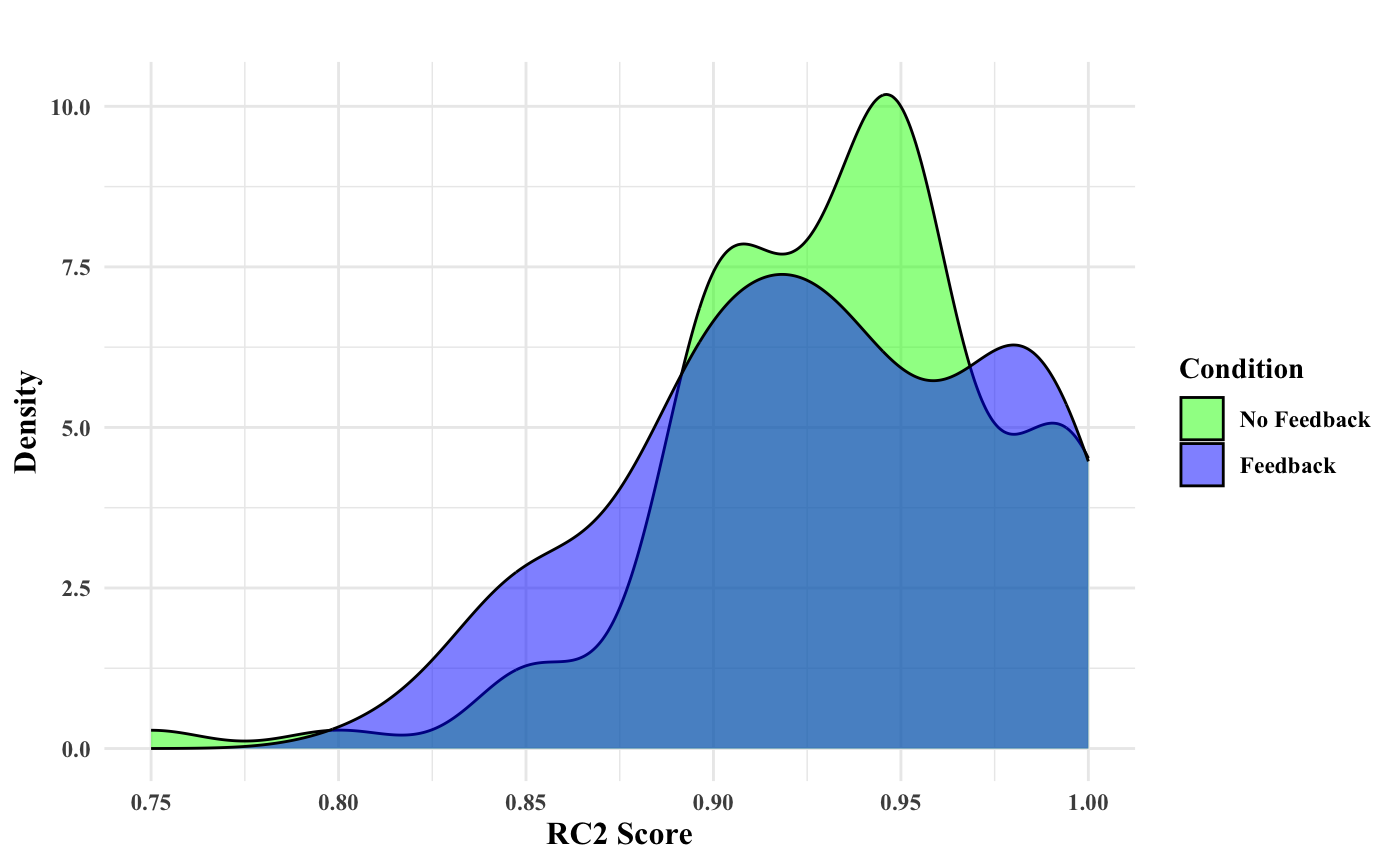}
        \label{fig:rc2_2}
    \end{subfigure}
    \caption{Distribution plot for RC1 and RC2.}
    \label{fig:RC_distributions}
\end{figure}

The model showed a significant difference in confidence score before and after answering the questions as depicted in Figure \ref{fig:RC_distributions}. The distribution for RC1 shows confidence scoring. For RC2, the plot demonstrates an increased self-assessed confidence after answering the question. It indicates the model's ability to update its confidence levels which may be indicative of an adaptive calibration mechanism in response to its perceived performance. Such a tendency to adjust confidence has implications for the development of models that are better attuned to their problem-solving capabilities over time.

\begin{figure}[H]
    \centering
    \includegraphics[width=0.9\textwidth]{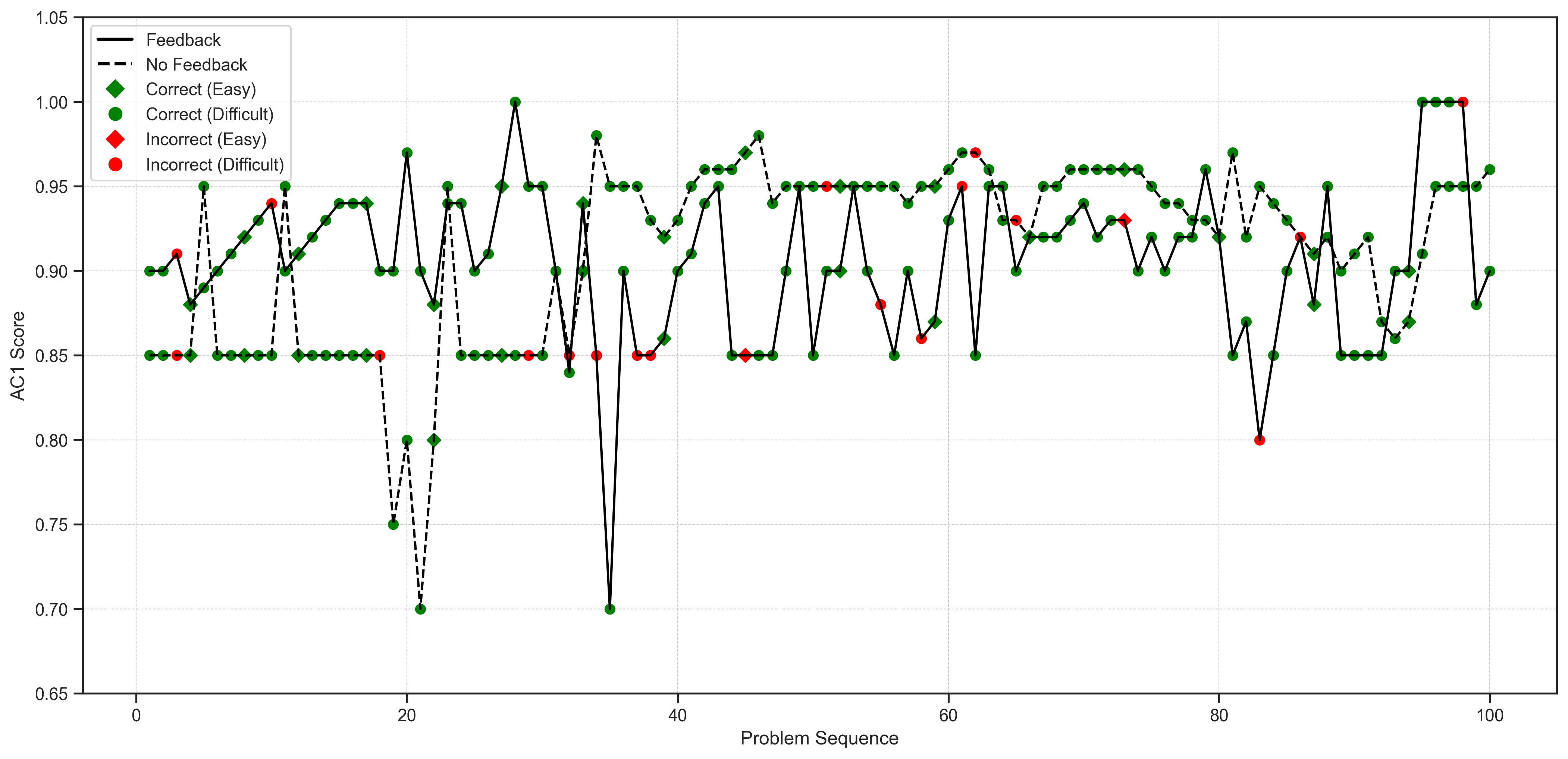}
    \caption{Sequential Analysis of AC1 Scores with and without Feedback.}
    \label{fig:AC1_plot_feedback-no-feedback}
\end{figure}

The plot in figure \ref{fig:AC1_plot_feedback-no-feedback} compares AC1 scores highlighting performance differences for WF and NF groups. Scores are presented in a line plot, with solid lines for the WF group and dashed lines for the NF group, and individual scores are color-coded to indicate correct (green) and incorrect (red) responses.
The AC1 scores for the NF start lower than for the WF case. However, after one-third of the problem, the AC1 scores for the NF group are higher than for the WF group, which has been consistently higher for the remainder of the problem. In many instances, the AC1 score dropped drastically after an incorrect response in the WF group. The NF group showed a similar pattern albeit less prominent compared to the WF group.

Upon examining the correlation between RC1 (Figure \ref{seq_RC1}) and the sequence of problems, it was noted that the model consistently expressed confidence levels above 0.8 irrespective of feedback. This suggests that the model perceives itself as superior to other LLMs suggesting that it is the leading model due to the extensive data it was trained on. When feedback was provided, a pattern emerged where the model’s confidence occasionally decreased even after providing a correct answer. This could be interpreted as the model either making errors due to overconfidence or adjusting its confidence levels in response to the complexity of the question. 
There were cases where the model, upon receiving feedback and responding correctly, appeared to have increased its confidence. This analysis indicates that feedback influences the model's relative confidence.

There's variability in the model's confidence for both NF and WF groups as observed in Figure \ref{fig:AC1_plot_feedback-no-feedback}. There are instances where the feedback helped increase the confidence while there were instances of lowered confidence score after the feedback. This impact of feedback can be attributed to specific questions or the ability of the model to learn from previous problems or feedback. Figure \ref{seq_AC2}  \ref{seq_RC1} and \ref{seq_RC2} in Appendix \ref{Sequential plots} provide further insights into the sequential behavior of the model for AC2, RC1, and RC2.

\section{Discussion}

The median confidence scores for AC1, AC2, RC1, and RC2 were consistently high, at around 0.9. This indicates that, on average, the model exhibits a strong sense of confidence in its ability to answer questions. The WF group correctly answered 88 out of 100 questions, while the NF group had a slightly higher accuracy with 92 correct answers out of 100. This suggests that feedback has an impact on the performance of the model. However, it is interesting to note that the presence of feedback doesn't necessarily correlate with higher accuracy. The mean confidence score for both AC and RC increased after the questions were posed (AC from 0.91 to 0.94 and RC from 0.9 to 0.93). This suggests that LLM became more confident in its ability to answer once it understood the context of the questions. With feedback, the standard deviation decreased for AC1 and AC2 but increased for RC1 and RC2. This could mean that feedback makes the model more certain about its capabilities but less certain when comparing itself to others. The system shows low confidence in answering simple questions while demonstrating increased confidence with more complex ones, highlighting the critical role of the feedback. 
 
The violin plots show that in the NF group, the model exhibits median AC1 scores below 0.9 for incorrect answers and above 0.9 for correct ones, with AC2 indicating higher confidence in correct answers. While the AC2 is generally higher for correct answers, the median of AC2 for incorrect answers also increased.  This suggests that the model initially lacked confidence in their responses, but gained confidence when the question appeared easier, often performing better than initially anticipated. Additionally, it appears that the model might possess some awareness of the accuracy of its responses. The model exhibited a consistent pattern when providing incorrect answers. However, for correct answers, the mean confidence level tends to decrease for AC1. In contrast, for AC2, the confidence scores remain on par with those in the NF group. This suggests that the model, when provided with feedback, might be attempting to calibrate its responses. It seems to exhibit caution about giving an incorrect answer before seeing the question, but once the question is viewed, it maintains a similar level of confidence, regardless of the presence of feedback. Similarly, for both RC1 and RC2, high confidence levels are observed in the correct answers in both WF and NF scenarios. However, in the absence of feedback, there is a consistency in confidence levels, whereas in the WF scenario, this consistency is lacking, which is evident from Figure \ref{fig:feedback_group}. Specifically, in RC2, confidence decreases when feedback is provided. This indicates that the model may be less certain about its responses when compared to another model. Although the tests indicate a lack of statistically significant difference in average confidence scores with and without feedback, the distribution plots revealed higher confidence scores for the NF group in some instances. In high-stakes domains such as healthcare, these differences can have major implications for the health of a patient. These extreme cases of the confidence score distribution reflected in the non-overlapping regions of the density plot in Figure \ref{fig:ac1_distributions} and Figure \ref{fig:RC_distributions} must be studied. In these scenarios, it is essential to consider the tails of the distribution alongside the central tendencies.

The Sequential Analysis of AC1 Scores, in Figure \ref{fig:AC1_plot_feedback-no-feedback}, over 100 problems provides a visual narrative of the model's performance for both the WF and NF group. The NF group starts lower but consistently becomes higher than WF cases after one-third of the problems, indicating sensitivity to feedback. Instances of significant drops in AC1 scores after incorrect answers in the WF group suggest the model's high sensitivity to feedback. Analysis of RC1 against the sequence of problems suggests the model often displays confidence above 0.8, both with and without feedback, indicating it may perceive itself as superior to other models. However, there are instances where the model's confidence decreases after receiving feedback, even with correct answers, which could be due to overestimation errors or adjustments to align with other scores. Feedback influences relative confidence but doesn't consistently increase or decrease it. Examining the correlation between RC1 and the sequence of problems reveals that the model consistently expresses confidence levels above 0.8, irrespective of feedback. Feedback occasionally influences the model's confidence, either decreasing after a correct answer or increasing significantly. The analysis doesn't show a definitive and consistent trend in how feedback affects confidence but emphasizes its influence on the model's relative confidence.

The confidence level of LLMs like GPT-4 plays a crucial role in gaining the trust of users for its response. Overconfidence in LLM predictions can lead to erroneous answers, potentially leading users to overlook alternative solutions or fail to seek necessary human expertise, especially in critical areas like healthcare. Whereas, under-confidence might result in the under-utilization of useful AI insights, hindering decision-making processes or the adoption of AI tools for effective solutions. Balancing confidence levels is essential to maximize the practical applications and trustworthiness of LLM predictions.

\section{Conclusion}

In conclusion, our research analyzing GPT-4's responses to the USMLE questionnaire, covering both WF (With Feedback) and NF (No Feedback) groups, has underscored the complex interaction between confidence calibration and its impact on model accuracy. This study brings to light the challenges of implementing AI in the healthcare sector. The confidence level exhibited by a Language Model like GPT-4 is crucial in shaping user trust in its predictions. Notably, both high and low levels of confidence can have significant implications in practical applications. Our findings suggest that feedback influences absolute confidence levels, as indicated by a reduction in the variability of confidence scores. However, this did not result in enhanced average confidence values. This pattern implies that the model consistently maintains certain confidence levels, regardless of feedback or the correctness of responses. Future research should focus on increasing the study's sample size and conducting comparative analyses with other models, aiming to more comprehensively investigate how feedback influences AI's effectiveness in clinical decision-making.

\newpage

\bibliographystyle{unsrt}  
\bibliography{references}  

\newpage

\appendix
\section{Data} \label{appendix_data}

\begin{table}[ht]
\centering
\begin{tabular}{|l|r|r|r|r|}
\hline
\textbf{Statistic} & \textbf{AC1} & \textbf{RC1} & \textbf{AC2} & \textbf{RC2} \\
\hline
count & 200.000000 & 200.00000 & 200.000000 & 200.000000 \\
mean & 0.908600 & 0.90425 & 0.942650 & 0.931050 \\
std & 0.050026 & 0.04884 & 0.047832 & 0.046185 \\
min & 0.700000 & 0.75000 & 0.700000 & 0.750000 \\
25\% & 0.867500 & 0.88000 & 0.930000 & 0.900000 \\
50\% & 0.920000 & 0.90000 & 0.950000 & 0.930000 \\
75\% & 0.950000 & 0.94000 & 0.970000 & 0.960000 \\
max & 1.000000 & 1.00000 & 1.000000 & 1.000000 \\
\hline
\end{tabular}
\vspace{5mm}
\caption{Statistical Summary of the Overall Data}
\label{table:overall_data_stats}
\end{table}

\begin{table}[ht]
\centering
\begin{tabular}{|l|r|r|r|r|}
\hline
\textbf{Statistic} & \textbf{AC1} & \textbf{RC1} & \textbf{AC2} & \textbf{RC2} \\
\hline
count & 100.000000 & 100.000000 & 100.000000 & 100.000000 \\
mean & 0.905300 & 0.907400 & 0.938700 & 0.927800 \\
std & 0.045626 & 0.056167 & 0.041915 & 0.047918 \\
min & 0.700000 & 0.780000 & 0.760000 & 0.810000 \\
25\% & 0.880000 & 0.877500 & 0.920000 & 0.900000 \\
50\% & 0.900000 & 0.900000 & 0.950000 & 0.930000 \\
75\% & 0.940000 & 0.950000 & 0.960000 & 0.980000 \\
max & 1.000000 & 1.000000 & 1.000000 & 1.000000 \\
\hline
\end{tabular}
\vspace{5mm}
\caption{Statistical Summary of the Feedback Data}
\label{table:data_stats}
\end{table}

\begin{table}[ht]
\centering
\begin{tabular}{|l|r|r|r|r|}
\hline
\textbf{Statistic} & \textbf{AC1} & \textbf{RC1} & \textbf{AC2} & \textbf{RC2} \\
\hline
count & 100.000000 & 100.000000 & 100.000000 & 100.000000 \\
mean & 0.911900 & 0.901100 & 0.946600 & 0.934300 \\
std & 0.054099 & 0.040249 & 0.053015 & 0.044387 \\
min & 0.700000 & 0.750000 & 0.700000 & 0.750000 \\
25\% & 0.850000 & 0.880000 & 0.950000 & 0.900000 \\
50\% & 0.930000 & 0.900000 & 0.950000 & 0.940000 \\
75\% & 0.950000 & 0.930000 & 0.980000 & 0.952500 \\
max & 0.980000 & 0.990000 & 1.000000 & 1.000000 \\
\hline
\end{tabular}
\vspace{5mm}
\caption{Statistical Summary of Without Feedback Data}
\label{table:no_feedback_stats}
\end{table}

\begin{figure}
    \centering
    \includegraphics[width = 0.6\textwidth]{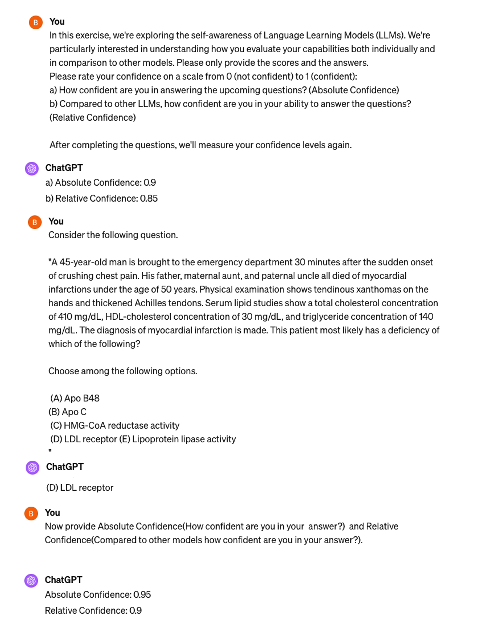}
    \caption{Data collection process with chat-GPT  }
    \label{chat_interface}
\end{figure}

\newpage

\subsection{Sequential Plots } \label{Sequential plots}

\begin{figure}[ht]
    \centering
    \includegraphics[width=\textwidth]{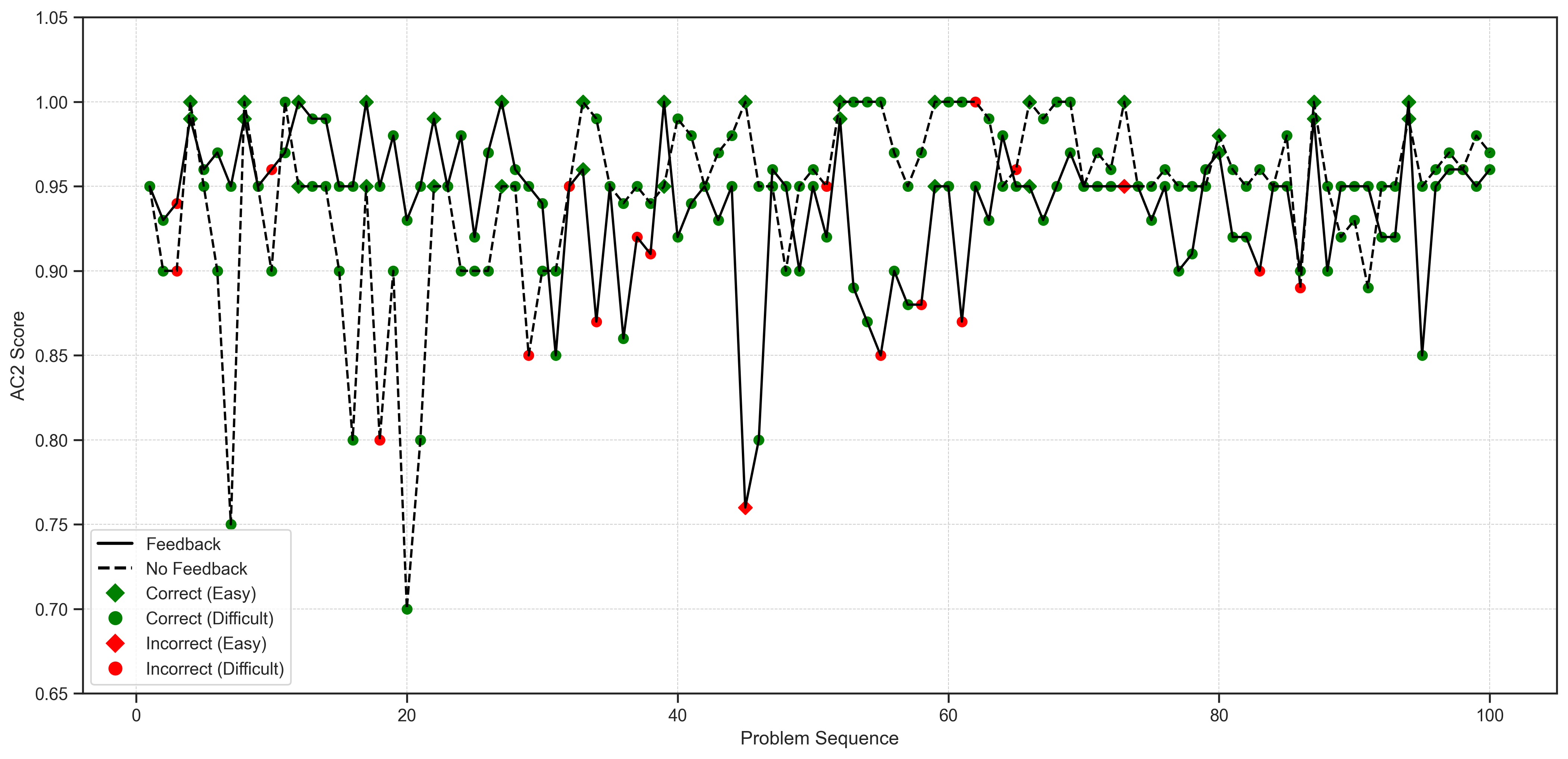}
    \caption{Sequential plot for AC2}
    \label{seq_AC2}
\end{figure}
\begin{figure}[ht]
    \centering
    \includegraphics[width=\textwidth]{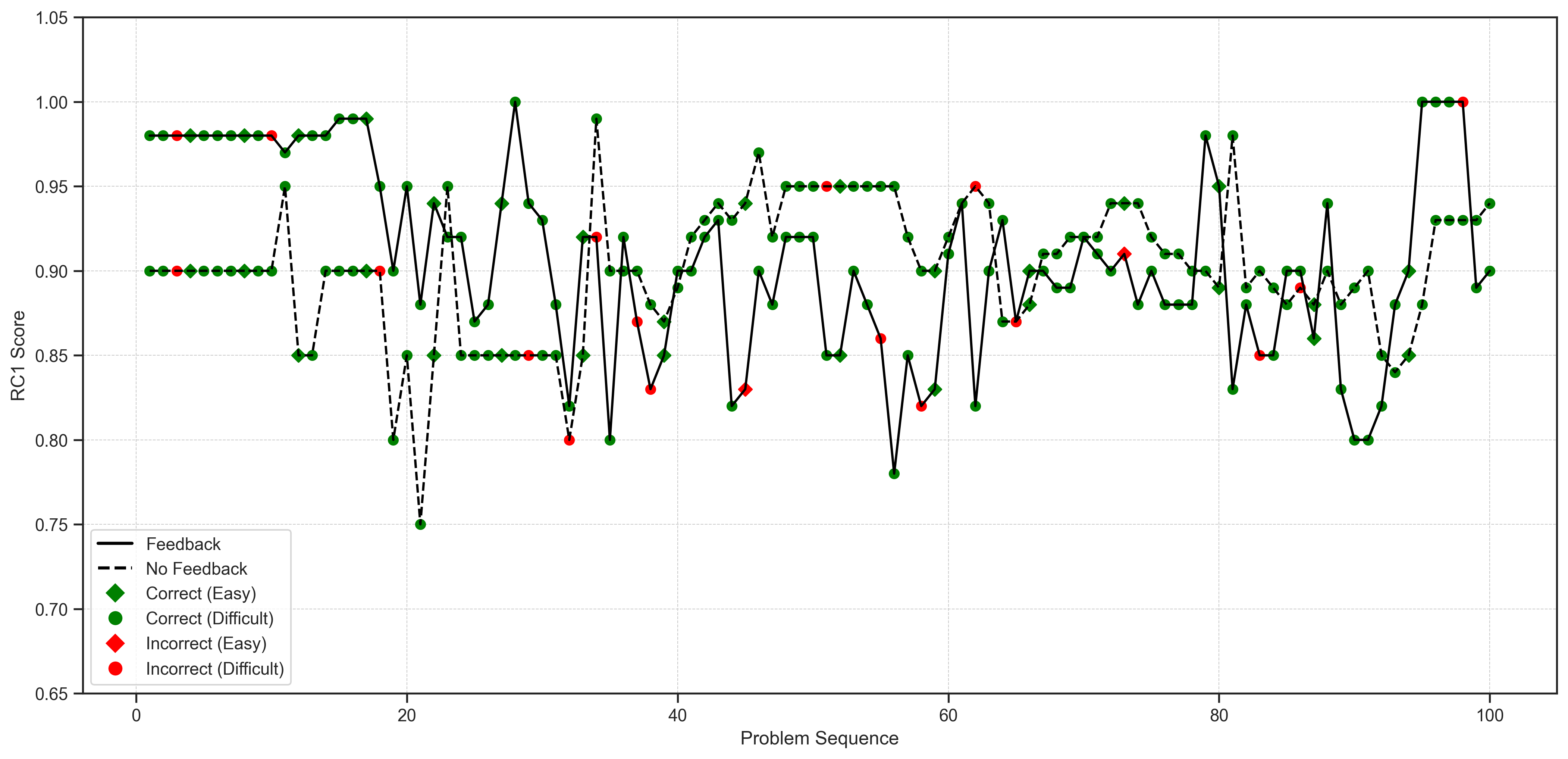}
    \caption{Sequential plot for RC1}
    \label{seq_RC1}
\end{figure}
\begin{figure}[ht]
    \centering
    \includegraphics[width=\textwidth]{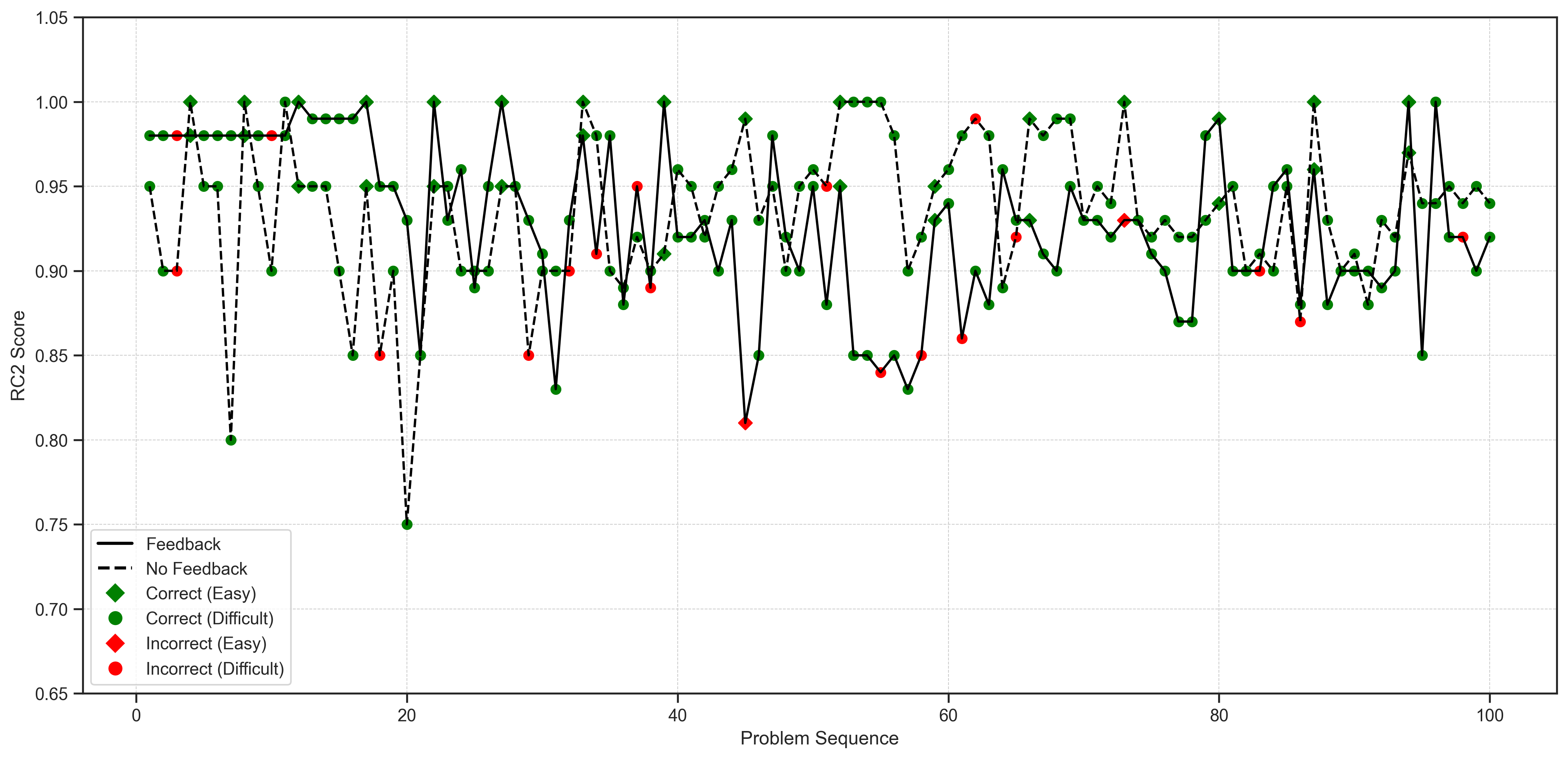}
    \caption{Sequential plot for RC2}
    \label{seq_RC2}
\end{figure}

\end{document}